\documentclass{article}

\usepackage[final]{neurips_2023} 

\usepackage[utf8]{inputenc} 
\usepackage[T1]{fontenc}    
\usepackage{hyperref}       
\usepackage{url}            
\usepackage{booktabs}       
\usepackage{amsfonts}       
\usepackage{nicefrac}       
\usepackage{microtype}      
\usepackage{xcolor}         

\usepackage{amsmath, amssymb}
\usepackage{graphicx}
\usepackage{enumitem}

\title{Neutralizing Structural Inequality in the Nigerian FinTech Sector}

\author{
  Muhammad Abdullahi Said
}

\begin{document}

\maketitle

\begin{abstract}
Algorithmic decision systems in financial services often rely on data proxies that inadvertently encode structural inequalities. This paper introduces a hierarchical human-AI triage model for Point of Sale fraud detection in the Nigerian FinTech sector. Adopting a We Are All Equal worldview, we address the challenge of discrimination laundering, wherein the system misinterprets infrastructure related aleatoric noise such as rural network timeouts as fraudulent intent. We implement a three-tier routing policy utilizing a calibrated ensemble model as a primary filter. The policy routes transactions characterized by epistemic uncertainty such as cold start new accounts to specialist analysts while reserving high stakes cases for a senior supervisor. To manage finite human capacity, we utilize a dynamic shadow price to ration human attention and implement a random audit mechanism to prevent human skill atrophy. Our experimental results demonstrate a statistically significant 1.88\% complementarity gap and a 24.79\% percentage point gain in fraud recall over an autonomous baseline. Crucially, the model reduces the regional performance gap from 19.43 to 2.88 percentage points, neutralizing structural bias. Hierarchical collaboration provides a robust mechanism for substantive equality of opportunity, ensuring that rural accounts are not excluded from the digital economy due to environmental brute luck.
\end{abstract}

\section{Motivation}
Point-of-Sale (POS) agents have emerged as the critical backbone of financial inclusion in Nigeria, acting as human bank branches for millions of unbanked citizens. Platforms like Moniepoint and OPay rely on these agents to process high volumes of cash-in and cash-out transactions. However, the rapid growth of this sector has brought a surge in sophisticated fraud, ranging from the use of stolen cards to complex money laundering schemes. To manage this at scale, financial institutions have deployed algorithmic decision systems (ADS) to flag suspicious behavior in real time.

While these models are efficient in urban centers like Lagos, their deployment in rural sectors reveals a significant socio-technical failure. Traditional AI models often rely on data proxies such as transaction velocity or failed retry rates to estimate fraud risk. In the Nigerian context, these proxies inadvertently encode structural bias. Rural agents frequently suffer from poor 3G connectivity, leading to frequent transaction timeouts and multiple retries. Under a standard ``What You See Is What You Get'' (WYSWYG) worldview, the AI interprets this infrastructure-related aleatoric noise as a sign of fraudulent intent. This results in ``discrimination laundering,'' where the brute luck of an agent's geographic location is converted into a low creditworthiness score, leading to automated rejection and permanent financial exclusion \cite{dearteaga2020case}.

In this paper, we propose a hierarchical human-AI triage model designed to navigate the fidelity-interpretability trade-off. Moving beyond formal equality, we adopt a \textit{We Are All Equal} (WAE) worldview to ensure substantive equality of opportunity for marginalized accounts. Our system implements a three-tier oversight regime: an autonomous AI filter for routine urban cases, a specialist analyst (Human 1) for infrastructure-related uncertainty, and a senior supervisor (Human 2) for high-stakes epistemic uncertainty. By calculating a dynamic shadow price to manage finite human capacity and utilizing random audits to prevent skill atrophy, we demonstrate that a collaborative team can neutralize structural bias while catching fraud that standalone models miss.

\section{System Design and Decision Space}
\begin{figure}[ht]
    \centering
\includegraphics[width=1.1\textwidth]{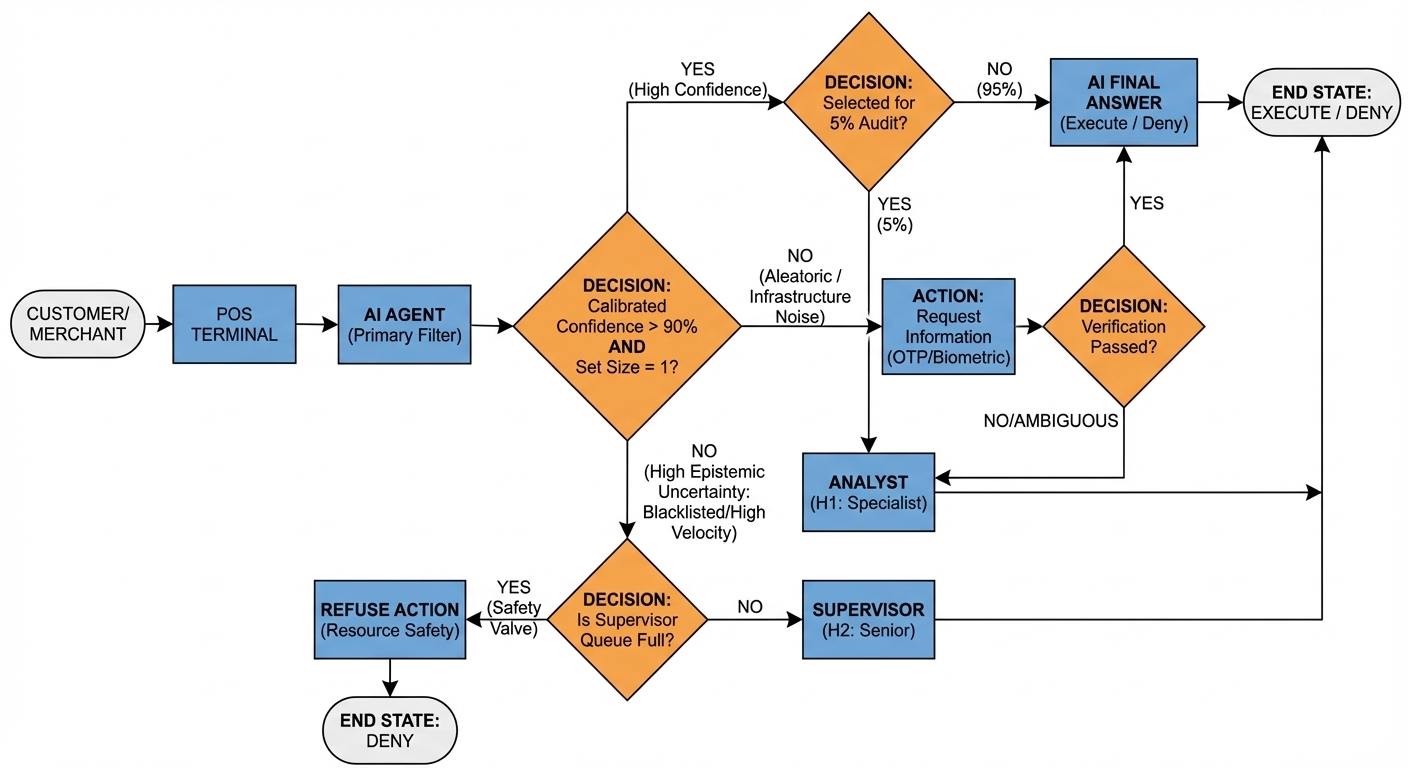} 
    \caption{\textbf{Hierarchical Routing Policy:} The triage workflow maps incoming POS transactions to the optimal agent. Epistemic uncertainty (blacklists) triggers Supervisor review, while aleatoric uncertainty (infrastructure noise) triggers Analyst review to prevent discrimination laundering.}
    \label{fig:routing_flowchart}
\end{figure}

Our system is modeled as a multi-level triage framework where (Figure \ref{fig:routing_flowchart}) a central routing policy $\pi$ maps each incoming transaction to the most appropriate agent. We define a context space $\mathcal{X}$ representing the features of each transaction, including the amount, geographic region, agent tenure, and transaction velocity.

\subsection{Agents and Action Space}
Following standard oversight taxonomy \cite{jorgensen2025documenting}, we define a three-tier hierarchy of decision makers, each with a unique expertise profile and associated cost. The \textbf{Autonomous AI} serves as the primary filter: it is computationally efficient and highly accurate on routine urban data, operating at near-zero cost, though it lacks the local context required for rural sectors. The \textbf{Human 1 Analyst} is a conditionally autonomous specialist who possesses high expertise in identifying rural infrastructure noise, intervening at a moderate escalation cost $C_{esc1}$. Finally, the \textbf{Human 2 Supervisor} is a human-led oversight agent reserved for the most complex cases; this senior supervisor provides high authority and acts as an oracle for legal and regulatory red lines, at the highest cost $C_{esc2}$ in the system.

Our policy $\pi$ selects from a five-fold action space $\mathcal{A}$ \cite{tang2026human}. In the \textbf{Answer} action, the AI or Human 1 approves or denies the transaction directly. \textbf{Escalation} routes an uncertain case from the AI to Human 1, while \textbf{Route to Specialist} bypasses routine channels to send high-stakes cases directly to Human 2. The \textbf{Request Information} action triggers an automated verification such as an OTP or biometric check before a decision is finalized. Finally, \textbf{Refuse} is a resource-dependent safety action: if a blacklisted account triggers a mandatory supervisor review but the Human 2 queue is at maximum capacity, the system issues an immediate refusal to prevent potential financial loss.

\subsection{Triple Pathway Routing Logic}
We implement a tripartite policy based on the nature of the uncertainty and the prevailing ethical worldview. \textbf{Pathway 1: The Utility Path} applies to low-value urban transactions with high model confidence, where the system maximizes beneficence by choosing the \textit{answer} action autonomously under a WYSWYG worldview. \textbf{Pathway 2: The Fairness Path} governs transactions from rural sectors, which often contain aleatoric noise due to network jitter. To prevent discrimination laundering, the system adopts a WAE worldview: when the AI is uncertain, it first issues a \textit{request information} action, and if ambiguity persists, the case is escalated to Human 1. \textbf{Pathway 3: The Red Line Path} handles cases involving blacklisted or new accounts, which carry high epistemic uncertainty. Adhering to the precautionary principle, these are routed directly to Human 2; if human-led oversight is unavailable, the system defaults to the \textit{refuse} action.

\section{Methodology}

\subsection{Uncertainty Quantification and Routing Signals}
To develop a robust routing policy $\pi$, we decompose model uncertainty into two distinct components. We first train a capacity-constrained Random Forest \cite{breiman2001random} ensemble ($M=100$) and apply Platt scaling for calibration. We validated calibration using the Weighted Expected Calibration Error (ECE), weighting the gap in each of the $B$ bins by the sample fraction \cite{naeini2015obtaining}:
\begin{equation}
ECE = \sum_{b=1}^{B} \frac{|I_b|}{n} |acc(I_b) - conf(I_b)|
\end{equation}

Our system identifies infrastructure-related aleatoric uncertainty ($u_{ale}$) by calculating the average predictive entropy across all individual decision trees $\theta_m$ in the ensemble:
\begin{equation}
u_{ale}(x) \approx \frac{1}{M} \sum_{m=1}^{M} H(Y | x, \theta_m)
\end{equation}
Transactions exceeding the 85th percentile of $u_{ale}$ are diagnosed as "network jitter" and routed via Pathway 2. To quantify epistemic uncertainty, we implement Adaptive Prediction Sets (APS). We calculate non-conformity scores $E_i$ based on the cumulative softmax mass required to reach the true label. The system generates a prediction set $C_{\alpha}(x)$ with a 90\% coverage guarantee; a Set Size $|C_{\alpha}(x)| > 1$ serves as the primary signal for model ignorance, triggering Pathway 3.

The optimal automation threshold $\tau^*$ is derived from the deployment cost structure, where the cost of missed fraud $C_{err}$ is weighted $10\times$ higher than human intervention $C_{esc}$:
\begin{equation}
\tau^* = 1 - \frac{C_{esc}}{C_{err}} = 0.90
\end{equation}

\subsection{Hierarchical Collaboration and Resource Management}
We model a three-tier hierarchy consisting of an Autonomous AI, a specialist Analyst (H1), and a senior Supervisor (H2). Our policy $\pi$ selects from a five-fold action space $\mathcal{A}$. To measure the value added by this collaboration, we calculate the Complementarity Gap ($\Delta_{comp}$) \cite{okati2021differentiable}:
\begin{equation}
\Delta_{comp} = \min(R_{model}, R_{human}) - R_{team}
\end{equation}
where $R_{human}$ is modeled as a varying expertise profile with high accuracy on rural noise but declining performance under high-velocity cognitive load.

To manage finite human capacity, we implement a dynamic shadow price $\lambda$ updated via dual ascent logic \cite{agrawal2014bandits}. As the H1 caseload approaches its 15\% budget, $\lambda$ increases, rationing human attention. Non-rural uncertain cases must clear the shadow price gate $(conf + \lambda) \geq \tau^*$. Critically, transactions exhibiting a rural geographic pattern bypass this gate under a \textit{We Are All Equal} (WAE) priority override to prevent "discrimination laundering." 

Finally, a Refuse action is implemented as a safety valve for Pathway 3; if a blacklisted account requires H2 review but the supervisor's 2\% budget is exhausted, the system prioritizes financial stability over autonomy and issues an immediate refusal.
\section{Experiments and Results}

\subsection{Experimental Setup and Dataset}
We utilized a 100,000 row stratified sample from the PaySim mobile money dataset \cite{lopez2016paysim}, modified for the Nigerian ecosystem. We augmented the fraud rate to 8.21\% and engineered features for geographic region (Urban vs. Rural) and account tenure. We utilized a three-way split (50\% train, 25\% calibration, 25\% test) to ensure statistical validity for both calibration and conformal prediction.

\subsection{Performance and Synergy}
The hierarchical system achieved an overall accuracy of 99.07\%, a 1.88\% improvement over the AI-alone baseline (97.19\%) (Table \ref{tab:performance}). As shown in the Risk-Coverage Curve (Figure \ref{fig:results_plots}, Top-Right) \cite{geifman2017selective}, the collaboration policy maintains significantly lower selective risk than the baseline at the optimal automation threshold of $\tau^* = 0.90$. Fraud recall increased from 66.39\% to 91.18\%, representing a massive 24.79\% gain. This synergy is validated by an overall complementarity gap $\Delta_{comp}$ of 1.880\% (Figure \ref{fig:results_plots}, Bottom-Right). The heatmap demonstrates that collaboration is effective across all quartiles for high-stakes "TRANSFER" and "CASH\_OUT" transactions. Furthermore, the Reliability Diagram (Figure \ref{fig:results_plots}, Bottom-Left) confirms the system's robustness with a near-perfect ECE of 0.0083, which ensures that the prediction sets used for routing are statistically valid.

\subsection{Fairness and Substantive Justice}
The triage logic successfully neutralized regional disparities encoded as infrastructure noise. According to the Accuracy by Transaction Type plot (Figure \ref{fig:results_plots}, Top-Left), the autonomous baseline suffered from a significant 19.43\% accuracy gap for rural-pattern transactions. By implementing a WAE-informed routing policy, our system reduced this gap to 2.88\%. Although rural transactions face higher latency (15.00 min vs 2.6 min), this inequality is ethically justified under a Rawlsian framework; it prevents "discrimination laundering" and ensures that rural agents are not excluded from the digital economy due to environmental brute luck.

\subsection{Operational Resource Management}
The AI autonomously handled 92.8\% of volume. Specialist Analyst H1 managed 6.7\% (primarily rural cases), and Senior Supervisor H2 handled 0.5\% (high-risk/blacklisted cases).

\section{Discussion and Conclusion}
By routing 6.7\% of transactions to Human 1, we achieved significant recall gains at the cost of 15.00 minutes of latency for rural agents. From a Rawlsian perspective, this inequality is ethically justified; it prevents discrimination laundering and ensures rural agents are not penalized for environmental brute luck. The 5\% random urban audit ensures Human 1 maintains a balanced expertise profile, preventing the analyst from becoming a specialist in noise alone \cite{dearteaga2020case}. The use of shadow price $\lambda$ suggests that human-AI collaboration is economically viable, reducing selective risk while requiring human review for only 7.2\% (Table \ref{tab:operations}) of transactions. The \textit{refuse} action for blacklisted accounts during bottlenecks prioritizes financial stability over individual autonomy. In emerging markets, algorithmic trust must recognize when a machine is out of its depth due to epistemic uncertainty.

In conclusion, this paper demonstrated that hierarchical human-AI triage can effectively neutralize structural bias in the Nigerian POS market. By moving from a WYSWYG to a WAE worldview, we closed a 19.43 percentage point accuracy gap. Successful collaboration requires a calibrated understanding of uncertainty and a deliberate allocation of finite human resources to ensure the digital economy remains inclusive and fair.

\newpage
\bibliographystyle{plain} 
\bibliography{references}

@inproceedings{geifman2017selective,
  title={Selective classification for deep neural networks},
  author={Geifman, Yonatan and El-Yaniv, Ran},
  booktitle={Advances in Neural Information Processing Systems},
  volume={30},
  year={2017}
}

@inproceedings{naeini2015obtaining,
  title={Obtaining Well Calibrated Probabilities Using Bayesian Binning},
  author={Naeini, Mahdi Pakdaman and Cooper, Gregory and Hauskrecht, Milos},
  booktitle={Proceedings of the AAAI Conference on Artificial Intelligence},
  volume={29},
  number={1},
  year={2015}
}

@article{okati2021differentiable,
  title={Differentiable learning under triage},
  author={Okati, Nastaran and De, Abir and Rodriguez, Manuel},
  journal={Advances in Neural Information Processing Systems},
  volume={34},
  pages={9140--9151},
  year={2021}
}

@inproceedings{jorgensen2025documenting,
  title={Documenting Deployment with Fabric: A Repository of Real-World AI Governance},
  author={Jorgensen, Mackenzie and Brogle, Kendall and Collins, Katherine M and Ibrahim, Lujain and Shah, Arina and Ivanovic, Petra and Broestl, Noah and Piles, Gabriel and Dongha, Paul and Abdulhussein, Hatim and others},
  booktitle={Proceedings of the AAAI/ACM Conference on AI, Ethics, and Society},
  volume={8},
  pages={1350--1362},
  year={2025}
}

@article{tang2026human,
  title={Human tool: An mcp-style framework for human-agent collaboration},
  author={Tang, Yuanrong and Peng, Huiling and Zhao, Bingxi and Ding, Hengyang and Song, Hanchao and Wang, Tianhong and Zhong, Chen and Gong, Jiangtao},
  journal={arXiv preprint arXiv:2602.12953},
  year={2026}
}

@inproceedings{agrawal2014bandits,
  title={Bandits with concave rewards and convex knapsacks},
  author={Agrawal, Shipra and Devanur, Nikhil R},
  booktitle={Proceedings of the fifteenth ACM conference on Economics and computation},
  pages={989--1006},
  year={2014}
}

@inproceedings{dearteaga2020case,
  title={A case for humans-in-the-loop: Decisions in the presence of erroneous algorithmic scores},
  author={De-Arteaga, Maria and Fogliato, Riccardo and Chouldechova, Alexandra},
  booktitle={Proceedings of the 2020 CHI conference on human factors in computing systems},
  pages={1--12},
  year={2020}
}

@article{breiman2001random,
  title={Random forests},
  author={Breiman, Leo},
  journal={Machine learning},
  volume={45},
  pages={5--32},
  year={2001},
  publisher={Springer}
}

@inproceedings{lopez2016paysim,
  title={{PaySim}: A financial mobile money simulator for fraud detection},
  author={Lopez-Rojas, Edgar Alonso and Elmir, Ahmed and Axelson, Stefan},
  booktitle={28th European Modeling and Simulation Symposium (EMSS)},
  pages={249--255},
  year={2016}
}

\newpage
\appendix
\section{Visualizations}

\begin{figure}[h]
    \centering
    \includegraphics[width=\textwidth]{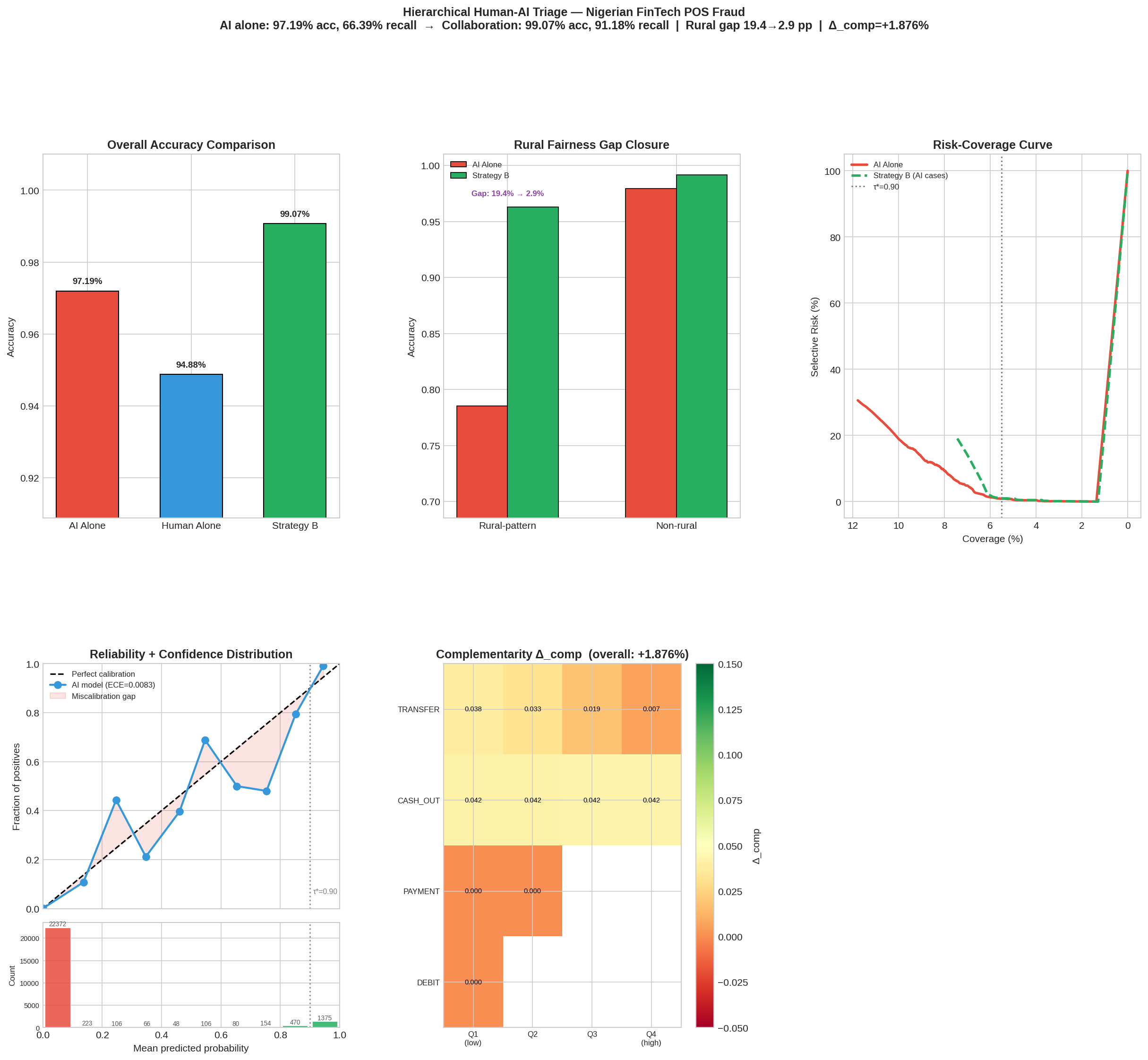} 
    \caption{Comprehensive Performance and Fairness Analysis. 
    Top-Left: Accuracy comparison showing the reduction of the rural performance gap from 19.43\% to 2.88\%. 
    Top-Right: Risk-Coverage curve demonstrating the collaboration policy's ability to maintain lower selective risk at higher coverage compared to the baseline. 
    Bottom-Left: Reliability diagram showing a highly calibrated base model (ECE=0.0083). 
    Bottom-Right: Complementarity Gap ($\Delta_{comp}$) heatmap showing an overall synergy gain of 1.880\%.}
    \label{fig:results_plots}
\end{figure}

\newpage
\section{Tables}

\begin{table}[h]
\centering
\caption{Performance and Synergy: AI Alone vs.\ Human-AI Collaboration}
\label{tab:performance}
\begin{tabular}{lccc}
\hline
\textbf{Metric} & \textbf{AI Alone} & \textbf{Strategy B} & \textbf{$\Delta$ (B $-$ AI)} \\
\hline
System Accuracy         & 97.19\% & 99.07\% & $\blacktriangle$ 1.88 pp  \\
Fraud Recall            & 66.39\% & 91.18\% & $\blacktriangle$ 24.79 pp \\
Rural Accuracy          & 78.52\% & 96.30\% & $\blacktriangle$ 17.78 pp \\
Non-Rural Accuracy      & 97.95\% & 99.18\% & $\blacktriangle$ 1.23 pp  \\
Rural--Non-Rural Gap    & 19.43 pp & 2.88 pp & $\blacktriangledown$ 16.55 pp \\
\hline
\end{tabular}
\end{table}

\begin{table}[h]
\centering
\caption{Operational Resource Allocation and Latency Under Strategy B}
\label{tab:operations}
\begin{tabular}{lcc}
\hline
\textbf{Agent / Metric} & \textbf{Caseload} & \textbf{Share of Volume} \\
\hline
Autonomous AI (Primary Filter) & 23,208 cases & 92.8\% \\
Specialist Analyst H1           & 1,669 cases  & 6.7\%  \\
Senior Supervisor H2            & 123 cases    & 0.5\%  \\
\hline
\textbf{Latency Metric} & \textbf{AI Alone} & \textbf{Strategy B} \\
\hline
Mean Wait Time   & 0.00 min & 3.08 min  \\
Rural Wait Time  & 0.00 min & 15.00 min \\
Urban Wait Time  & 0.00 min & 2.60 min  \\
\hline
\end{tabular}
\end{table}

\end{document}